%% file: ms.tex
\RequirePackage{filecontents}

\documentclass{elektr}
\usepackage[all]{xy,xypic}
\usepackage{natbib}
\usepackage[utf8]{inputenc}
\usepackage[english]{babel}
\usepackage[dvipsnames]{xcolor}
\usepackage[pagewise]{lineno}
\usepackage{hyperref}
\usepackage{amsfonts,amssymb,amsgen,amsopn,amsbsy,theorem,graphicx,epsfig}
\usepackage{eufrak,amscd,bezier,latexsym,mathrsfs,eurosym,enumerate}
\usepackage{comment}
\usepackage{subcaption}
\usepackage{adjustbox}
\usepackage[utf8]{inputenc}
\usepackage[english]{babel}
\usepackage{algorithm}
\usepackage{algpseudocode}
\usepackage{courier}
\usepackage[dvipsnames]{xcolor}
\usepackage[pagewise]{lineno}
\usepackage{relsize} 
\usepackage{graphicx} 
\usepackage{placeins}
\usepackage{float}
\usepackage{subcaption}
\usepackage{array,booktabs}
\usepackage{tabularx}
\usepackage[normalem]{ulem}
\usepackage{color}
\usepackage{multirow}
\usepackage{fancyhdr}
\usepackage[symbol]{footmisc} 
\usepackage{amsmath,amssymb,enumitem}
\usepackage{perpage} %the perpage package
\usepackage{comment}
\usepackage{breqn}
\usepackage{relsize} 
\usepackage{placeins}
\usepackage{breqn}
\usepackage{lineno}
\usepackage{titlesec}
\usepackage{hyperref}
\usepackage{datetime}
\hypersetup{
	colorlinks=true,
	urlcolor=blue,
	citecolor=blue}
\usepackage{cleveref,multirow}

\date{\today}
\newcommand*{\bdbackslash}{\textunderscore\kern-.1355ex\textunderscore}
\newcommand*{\cdbackslash}{\_\kern-.1355ex\_}

\title{Self-supervised Vector-Quantization in Visual SLAM using Deep Convolutional Autoencoders \vspace{1cm} \\  \today
}
\author[ Amir Zarringhalam]{
	\textbf{ Amir Zarringhalam$^{1}$\thanks{am.zarringhalam@aut.ac.com}~,Saeed Shiry Ghidary$^{2}$, Ali Mohades Khorasani$^{3}$} \vspace{1cm} \\
	
	$^{1}$Amirkabir University of Technology, Computer Science and  Mathematics, 424 Hafez Ave, Tehran, Iran\\
	$^{2}$Staffordshire University, School of Digital, Technologies and Arts, College Rd, Stoke-on-Trent ST4 2DE, United Kingdom,\\
	 Amirkabir University of Technology, Computer Science and  Mathematics, 424  HafezAve, Tehran, Iran\\
	$^{3}$Amirkabir University of Technology, Computer Science and  Mathematics, 424  HafezAve, Tehran, Iran
	\\ [1.8em]
}

\input{elksty.tex}

\begin{document}
\maketitle

\begin{abstract}
	In this paper, we introduce AE-FABMAP, a new self-supervised bag of words-based SLAM method. We also present AE-ORB-SLAM, a modified version of the current state of the art BoW-based path planning algorithm. That is, we have used a deep convolutional autoencoder to find loop closures. In the context of bag of words visual SLAM,
	vector quantization (VQ) is considered as the most time-consuming part of the SLAM procedure, 
	which is usually performed in the offline phase of the SLAM algorithm using unsupervised algorithms such as Kmeans++. 
	We have addressed the loop closure detection part of the BoW-based SLAM methods in a self-supervised manner, by integrating an autoencoder for doing vector quantization. This approach can increase the accuracy of large-scale SLAM, where plenty of unlabeled data is available. 
	The main advantage of using a self-supervised is that it can help reducing the amount of labeling. 
	Furthermore, experiments show that autoencoders are far more efficient than semi-supervised methods like graph convolutional neural networks, in terms of speed and memory  consumption. 
	We integrated this method into the state of the art long range appearance based  visual bag of word SLAM, FABMAP2, also in ORB-SLAM. Experiments demonstrate the superiority of this approach in indoor and outdoor datasets over regular FABMAP2 in all cases, and it achieves higher accuracy in loop closure detection and trajectory generation.
	%\keywords{Electronics, instructions for authors, manuscript template}	
\end{abstract}

\section{Introduction}
Humans can benefit from autonomous vehicles and robots that operate long term in unaffordable circumstances, such as underground operations and navigating on the surface of other planets. Current autonomous navigation systems heavily depend on the Global Positioning System (GPS) for localization and navigation, which may not be available in many situations, such as  populated urban areas,  where skyscrapers stand and obscure the observable satellites. Also in indoor environments, GPS 
does not work reliably, since it depends on open sky factor. 

Even when we are not limited by these factors, GPS does not  provide us an accurate enough localization. In such situations, robots can use visual SLAM for the purpose of navigation. In the last four decades, researchers have proposed many solutions to the terrestrial visual SLAM problem, which most of them  use Kalman filters and sparse information filters to solve the SLAM problem. However, they can not produce large-scale maps, essentially due to the higher computational costs and uncertainties. For large-scale navigation, vSLAM plays a key role, however, there are mainly two issues happening while using the SLAM algorithm for large-scale scenarios. First, localization and positioning tend to drift for  autonomous vehicles that are planing to drive over hundreds of kilometers in different air conditions. Second, maps do not necessarily remain viable under different driving conditions, and the positioning tends to deviate from the truth trajectory as the driving distance increases. Therefore, without prior knowledge of the environment, it is almost impossible to ensure correct localization of  the vehicle over several kilometers. This results in turning our attention toward maps with sufficient localization accuracy.    Several papers \citep{Paz468056}, \citep{Se2005VisionbasedGL} and \citep{1492475} have addressed this problem by creating  sub-map and multiple maps. However these methods impose proper data association assumption, to build large-scale maps that need consistent improvement. Therefore, these methods do not have a  solution for large-scale implementation for increasingly unstructured environments.   In this paper, we attempt to address the  loop closure detection (LCD) problem, as the core of the vSLAM method,  which is prone to many errors such as  perceptual aliasing. \\

\section{Related Works}
Methods that have solved large scale loop closure detection using bag of words schema, up to now, are usually unsupervised, like SGNNS \citep{DBLP:journals/corr/HajebiZ13} which is the state of the art (in terms of speed) in this scope. Considering features extracted from sequence of images form a graph, SGNNS  starts from a random node, while imposing a threshold on the nearest neighbor to that node, it incorporates KNN search and clusters the features. The main advantage of SGNNS over linear search and Kmeans, is its higher speed. Nevertheless, the constraint induced by this method for the nearest neighbor distance is inevitable. As another example Nishant, etal \citep{KEJRIWAL201655} proposed a bag of words pair (BoWP) based loop closure detection method that resolves the simple
BoW method limitations like perceptual aliasing. This is achieved through incorporating the spatial occurrence of words in vector quantization procedure. They use a Bayesian probabilistic schema for loop closures detection. 
\\
As another example, iBoWLCD (incremental BoW loop closure detection) \citep{DBLP:journals/corr/abs-1802-05909}, proposes a novel appearance based model for loop closure detection. This method uses incremental BoW schema on binary features. It  generates an on-line BoW without deleting any visual words learned during training. \\
In another work, Dorian Glovez \citep{GlvezLpez2012BagsOB} uses hierarchical BoW schema to detect revisited places. In order to build the hierarchical bag of words, which is structured as a tree, a set of binary features are first discretized to set of k cluster centroids, therefore a more compact representation of the features is obtained. The set of clusters constructed in this stage form the first level of the tree. This procedure is repeated on every cluster to make the subsequent levels of the tree.
\\

To mention other examples, in \citep{IB} an appearance based method for loop closure detection, IBuILD (Incremental Bag of Binary Words for Appearance Based Loop
Closure Detection), is presented. This approach focuses on building  incremental binary vocabularies. This on-line method does not require  learning vocabularies in every iteration of the SLAM algorithm, it only depends on scene appearance for loop closure detection. It also doesn't need GPS estimation and odometry information, because it builds vocabulary based on tracking the features between two consecutive images, to impose pose invariance assumption. Furthermore,  this approach uses simple likelihood function to  generate suitable loop closure detection candidates. It also uses provisional consistency constraint to filter non-homogeneous closures out.\\

In another work \citep{6094820}, a method for pose graph optimization, and increasing performance of loop closure detection is presented. This is achieved using a combination of two novel metrics, expected information gain and visual word saliency. This paper uses global and local saliency to measure scarcity of an image in a dataset and en-richness of the candidate images for loop closure detection. The later measure is defined using bag of visual words histogram entropy for key-frames.\\

In PTAM (Parallel Tracking And Mapping) \citep{PIRE201727}, that ORBSLAM is heavily based on, a camera pose detection method in an unknown scene is introduced. Although this issue has previously been well studied, PTAM introduces a parallelism, a method for tracking and mapping in small scale augmented reality framework. Specifically, tracking and mapping are attempted in parallel using two threads. One is in charge of tracking  the cluttered hand movements, while other builds 3D map using extracted features from previously  observed frames. This allows us to perform costly batch optimization operation in real time, which results in maps with thousands of land marks that can be tracked per frame  with sufficient accuracy.

\section{Methods}

In this section, we briefly review the basic methods used in this paper. Our main modifications to these methods are also presented. First, we review the  deep convolutional autoencoder (AE) used in  this work. Next, we briefly review the  FABMAP2 formulation, and  introduce our novel method AE-FABMAP2  which integrates an AE for vector quantization. Then, we review ORB-SLAM and our modified version,    AE-ORB-SLAM. We have also integrated an autoencoder on BoW SLAM which we explain at the end of this section.
\subsection{Deep Convolutional Autoencoder (AE)}
CAE is a model based  on the  encoder-decoder paradigm. First the encoder transforms the input into a lower dimensional space, then a decoder is regulated to reconstruct  initial input from the low dimensional representation.  This is achieved by minimizing of the cross entropy cost function \citep{DBLP:journals/corr/TurchenkoCL17}: \\

\begin{equation}
	\label{equation1}
	E = \mathlarger{\frac{1}{N} \sum_{n=1}^{N}(y log(\hat{y_n}) + (1-y_{n})log(1 - \hat{y_n}))} 
\end{equation}

CAE uses convolution/deconvolution layers  for the  encoding/decoding part. These layers are followed by an activation function, and they are described as follows:\\
\begin{equation}
	\label{equation2}
	h^{k} = f(\mathlarger{ \sum_{l \in L} x^{l}  \circledast w^{k} +b^{k}})
\end{equation}

In Formula \ref{equation2}, $k$ is the latent representation of the $k^{th}$ feature map of the current layer, $f$ is a non-linear activation function. $\circledast$ denotes the 1D convolution operation,
$w^{k}$ and $b^{k}$ are the weights (filters) and bias of the $k-$th feature map of the current layer, respectively.
The CAE layers are arranged as follows:\\
\textbf{Encoder:}
\begin{itemize}
	\item First layer consists of $32-3 \times 1$ filters followed by a max-pooling layer. 
	\item The second layer will have $64-3 \times 1$ filters followed by a down-sampling layer. 
	\item Final layer consists of $128-3 \times 1$ filters. 
\end{itemize}
\textbf{Decoder:}
\begin{itemize}
	\item First layer consists of $128-3 \times 1$ filters followed by a up-sampling layer. 
	\item The second layer will have $64-3 \times 1$ filters followed by another up-sampling layer. 
	\item Final layer consists of $1-3 \times 1$ filters. 
\end{itemize}
\subsection{Self-supervised Vector Quantization(SVQ)}
\begin{linenomath}
	Vector quantization is a sampling mechanism, which gets an input from elements of a vector space, and returns a set of indicies denoted with $ i $, as the result. The results are also coming from the same vector space. Here $ i $ refers to clustering centroids, which is a countable set, $i \in \{1,..,|\mathcal{C}| \}$, where $|\mathcal{C}|$ is the number of  clustering that we get from applying a convolutional  deep Autoencoder on the extracted SURF feature of the images. To be more specific, suppose $ D $ is the set of descriptors of a particular image in the database then:
	\begin{equation}
		\label{equation3}
		\hat{Q(D)}_{i} = \{j: \Vert D[i,] -C[j,] \Vert_{2} \leqslant \Vert D[i,] -C[k,] \Vert_{2} \forall k  \in 
		\{1, .. , |\mathcal{C}|\} \}\\
	\end{equation}
	is the quantized representation of features in that particular image.
\end{linenomath}
\subsection{FAB-MAP2}
\begin{linenomath}
	Here, we give a brief overview of the FAB-MAP2 algorithm. A detailed explanation of the algorithm can be found in\citep{newman07isrr}. FABMAP2 is a generative model which uses the following fact in the construction of BoW representation  of images: Words that co-occur together usually originate from the same place. The configuration mentioned about FABMA2P, enables this method to detect places having too many features in common.
	In this method, the observation $Z$ of an image at time $k$ is reduced to some binary vectors. 
	That is $Z_{k} = \{z_{1},...,z_{|\mathcal{C}|}\}$, where $z_{i}$ represents word $i$ in the  vocabulary, and  indicates presence or absence of word $i$ in the vocabulary. The vocabulary in FAB-MAP2 is obtained trough clustering the extracted SURF features of the images using Kmeans. The overall observations up to time $k$ is denoted by  $Z^{k}$. The map of the environment is constructed from  locations  $\mathcal{L}^{k} = \{L_{1},...,L_{n_{k}}\}$. Each of these locations is mapped to an appearance model, for example $L_{i}= \{p(e_{i}=1|L_{i}),....,p(e_{|\mathcal{C}|}=1|L_{i})\}$.
	\begin{comment}
		in which $c's$ are scene's location that are constructed independently from situations. A detector model in this system is defined as :\\
		\begin{center}
			$ \mathcal{D} = 
			\left\{
			\begin{array}{ll}
				p(z_{q}=1|e_{q}=0) \hspace{0.5cm} False \hspace{0.3cm} positive \hspace{0.3cm} probability \\
				p(z_{q}=0|e_{q}=1) \hspace{0.5cm} False \hspace{0.3cm} negative \hspace{0.3cm} probability 
			\end{array}
			\right.
			$
		\end{center}
		the purpose of introducing scene elements is to construct a frame work, to incorporate multiple sources data, each of which having it's own error function. In addition it helps us to factor $p(z|L_i)$ out, into two terms.
	\end{comment} 
	Suppose the robot is in the middle of the way and it has constructed a partial map of the environment. As the robot acquires  new observation the likelihood of being in each of these locations is calculated, considering  the observation up to the time i.e, $p(L_i|Z^{k})$ is known. This term is obtained from the following formula:\\
	\begin{equation}
		\label{equation4}
		\centering
		P(L_{i}|\mathcal{Z}^{k}) =\mathlarger{\frac{p(Z_{k}|L_{i},\mathcal{Z}^{k-1})p(L_{i}|\mathcal{Z}^{k-1})}{p(Z_{k}|\mathcal{Z}^{k-1})}} 
	\end{equation}
	In  formula \ref{equation4},  $p(L_{i}|\mathcal{Z}^{k-1})$  is the prior probability of the robot location, and $p(Z_{k}|L_{i},\mathcal{Z}^{k-1})$ is the observation likelihood.\\
	To simplify the  evaluation  of  $p(Z_{k}|L_{i},\mathcal{Z}^{k-1})$, we suppose that observations in current time and the previous time are independent. 
	\begin{comment}
		Therefore we have
		$p(Z_{k}|L_{i}) = p(z_{n}|z_{1},...,z_{n-1},L_{i}) \times p(z_{n-1}|z_{1},...,z_{n-2},L_{i}) \times p(z_{2}|z_{1},L_{i}) \times p(z_{1}|L_{i})$. \\
	\end{comment}
	%Due to conditional dependency between words, calculating  $p(Z_{k}|L_{i})$ is intractable. 
	Therefore using the naive Bayes formula we will have:\\
	\begin{equation}
		\label{equation5}
		p(Z_{k}|L_{i}) \approx p(z_{r}|L_{i}) \mathlarger{ \prod_{q=1}^{|\mathcal{C}|} p(z_{q}|z_{P_q},L_{i})} 
	\end{equation}
	\begin{comment}
		The term $p(z_{q}|L_{i})$ can be expanded as: \\
		
		\begin{center}
			$p(z_{q}|L_{i}) = \sum_{s \in \{0,1\}} p(z_{q}|e_{q}=s,L_{i})p(e_{q}=s|L_{i})$.
		\end{center} 
		
		Considering the errors to be independent from locations we can write:\\
		
		$p(z_{q}|e_{q},L_{i}) = p(z_{q}|e_{q}) \rightarrow p(z_{q}|L_{i}) = \sum_{s \in \{0,1\}} p(z_{q}|e_{q} =s)p(e_{q}=s|L_{i})$ \\
		
		And in order to make it tractable, we assume\\
		
		$p(e_{q}=S_{s_{q}}|z_{p_{q}},L_{i}) = p(e_{q}=s_{e_{q}}|L_{i})$,\\
		which results in :
		$p(z_{q}|z_{p_{q}},L_{i}) = \sum_{s_{e_{q}} \in \{0,1\}}p(z_{q}|e_{q}=s_{e_{q}},z_{p_{q}})p(e_{q}=s_{e_{q}}|L_{i})$.\\
	\end{comment}
	In order to calculate the probabilities of absence of the observations in the map, it is needed to calculate 
	$p(z^{k}|z^{k-1})$, to divide the map space into two parts: mapped and unmapped.
\end{linenomath}
\begin{equation}
	\label{equation6}
	p(Z^{k}|Z^{k-1}) = \sum_{m \in \mathcal{L}^{k}}p(Z^{k}|L_{m})p(L_{m}|Z_{k-1}) + \sum_{u \in \mathcal{\overline{L}}^{k}}p(Z^{k}|L_{u})p(L_{u}|Z_{k-1})
\end{equation}

In  formula \ref{equation6}, we can't compute the second term. Because it contains unvisited environments, though  we can use mean field approximation to obtain this term, i.e: 
$p(Z_{k}|L_{avg}) =  \sum_{u \in \overline{\mathcal{L}^{k}}} p(L_{u}|Z_{k-1})$,
where $\sum_{u \in \overline{\mathcal{L}^{k}}} p(L_{u}|Z_{k-1})$
is the posterior probability of being a new place. \\
\subsection{AE-FABMAP}
In AE-FABMAP we have replaced the common unsupervised method Kmeans, with an autoencoder to quantize the image features. That is, after extracting the features from the image database, we feed them into a deep, 19 layer  convolutional  autoencoder to obtain the labels. Then, the cluster centroids and the BoW representation of the images are created using the obtained labels. Then, we fuse the BoW representation and cluster centroids obtained in the previous step into the FABMAP2 method to find the similarity between images in form of a confusion matrix. The general description of the method AE-FAMBAP is presented in Algorithm \ref{alg}.

\begin{algorithm}
	\caption{AE-FABMAP  \newline Input: Training and Testing Images \newline Output: Confusion Matrix} 
	\label{alg}
	\begin{algorithmic}[1]
		\Procedure{AE-FABMAP}{Train Images, Test Images}
		\State  Extract 128-d SURF features from all collected images 
		\State  Both for train and test dataset
		\State  Apply PCA on the train and test images matrix
		\State  Obtain BoW representation of train \& test Datasets using an AutoEncoder
		\State  Train a Chow-Liu tree (from training Dataset)
		\State  i = Number of images in test dataset
		\While{i $>$ 0}	
		\State Pick the row corresponding to the $i^{th}$ image in the BoW representation   
		\State Use a motion model calculate loop closure detection probability
		\If {LCD criteria is met}
		\State	i - -			
		\State	Continue
		\Else 
		\State Add new location
		\EndIf	
		\State	i - -				
		\EndWhile
		\State  Return confusion matrix
		\EndProcedure
	\end{algorithmic}
\end{algorithm}

The general procedure of AE-FABMAP is described  in algorithm \ref{alg}. In this algorithm,  part of LCD criteria is as follows: If the current observation matches a location
with probability higher than a user specified threshold (e.g., p $>$ 0.999), we associate the observation
with that location \cite{5613942}.\\
%Specifically, AE-FABMAP fills the confusion matrix  using  formulas  \ref{equation7}  and \ref{equation8}:\newline

A more detailed version of the algorithm \ref{alg} is described in the algorithm \ref{loop} and it's continuation in algorithm \ref{loop1}. In the first part of the  algorithm \ref{loop}, the default likelihood is calculated once (When a location is added to the map for the first time), that is we initialize the the locations default likelihood with $d_1$. Here we assume a null observation with $ z_q = 0 $ for every $q$, and it is equal to the summation of $ D_q $'s that is calculated for every word in the location.
We also notice that the default likelihood is different for every location.\\

While processing new observation, the default likelihood of a place is adjusted using the variables $d_2, d_3$ and $d_4$. In the next part of the algorithm \ref{loop} the observation likelihood are adjusted according to the content of current observation. That is, we consider the  locations where $q$ are observed in. All the log-likelihood of all locations are subtracted from the default value and they are updated with the proper values $d_3$ or $d_4$.
In the last part of the algorithm we consider the words, in which the observation occurs in the parent node of Chow-Liu tree but not the node itself, the log-likelihoods are adjusted with $q_2$.

\begin{algorithm}[!htp] 
	\caption{}
	\label{loop}
	\begin{algorithmic}[1]	
		\State Vector $<$Imatch$> $ $\&$ matches
		\State Vector test$\_$Image$\_$Descriptor
		\For{$i = 0$ ; $\#$ test $\_$images ; i++} 
		\State Vector $<$Imatch$> $  QueryMatch
		\State Vector $<$Double$> $ Default
		\State map $<$int, Vector $<$int$>$ $>$ inverted$\_$map;
		%\If {queryImgDescriptor.at$<$float$>$ (0,q) $>$ 0}
		\For{$k = 0$ ; $\#$ test $\_$images ; k++}
		\State Default.push$\_$back(0);
		\For{$q = 0$ ; q $< $ CLtree(0,q) ; q++}
		\If{test$\_$images(k)(0,q) $>$ 0}
		\State {defaults.[end] += d1[q]} 
		\State {inverted$\_$map[q].append(defaults.size())}
		\EndIf
		\EndFor	 
		\State Vector::iterator Lwidth, child;
		\State Vector$<$Double$>$ liklihoods = Defaults;
		\For{$q1 = 0$ ; q1 $< $ CLtree.cols() ; q1++}
		\If{queryImgDescriptor(0,q1) $>$ 0}
		\For{Lwidth=inverted$\_$map(q1).begin();Lwidth!=inverted$\_$map(q1).end(); Lwidth++}
		\newline
		\If{queryImgDescriptor(0,$P_q(q1)$) $>$ 0}
		\State liklihoods[*Lwidth]+= $d_4[q1]$
		\Else	
		\State liklihoods[*Lwidth]+= $d_3[q1]$
		\EndIf
		\For{child=children(q1).begin();child!=children(q1).end(); child++}
		\If{queryImgDescriptor(0,*child) $==$ 0}
		\For{Lwidth=inverted$\_$map(*child).begin(); \\
			Lwidth!=inverted$\_$map(*child).end(); Lwidth++} 
		\State liklihoods[*Lwidth]+= $d_2[*child]$;\\
		\EndFor
		\EndIf
		\EndFor
		\EndFor
		%\EndFor
		\EndIf
		\EndFor
		\EndFor
		\For{ int k=0 ; liklihood.size() ; k++}
		\State QueryMatch.pushback(Imatch(0,k,liklihood(k),0));
		\EndFor
		\State continued...
		\EndFor   		  
	\end{algorithmic}
\end{algorithm}

\begin{algorithm}[!htp] 
	\caption{Continued}
	\label{loop1}
	\begin{algorithmic}[1]
		\For{ int j=1 ; QueryMatch.size() ; j++}
		\State QueryMatch[j].QueryIndx = i;
		\EndFor
		\State matches.insert(matches.end(), QueryMatch.begin(), QueryMatch.end());
	\end{algorithmic}
\end{algorithm}
\newpage
\begin{flalign}	
	\label{equ7}
	d2_{num} = \frac{CLtree(1,q) \times 0.61 \times (1-CLtree(2,q))}{(1-CLtree(1,q)) \times 0.39 \times CLtree(2,q)} &&
\end{flalign}
\begin{flalign}
	d_2 = \log(\frac{d2_{num}}{1- d2_{den}}) - d_q &&
\end{flalign}
\begin{flalign}
	d2_{den} =\frac{CLtree(1,q) \times 0.61 \times CLtree(2,q)^{2} \times 0.39}{[(1-CLtree(1,q)) \times 0.61 \times (1-CLtree(2,q)) + CLtree(1,q) \times 0.39 \times CLtree(2,q)] \times (1- 0.39\times CLtree(1,q)) }  &&
\end{flalign}
\begin{flalign}
	d3_{den}= \frac{0.61 \times CLtree(1,q) \times (1-CLtree(1,q)) \times 0.39 \times CLtree(3,q) }{(1-CLtree(1,q)) \times CLtree(1,q) \times 0.61 \times (1-CLtree(3,q))} &&
\end{flalign}
\begin{flalign}
	d3_{num} = \frac{(1-CLtree(1,q)) \times 0.39 \times CLtree(3,q)}{(1-CLtree(1,q)) \times 0.39 \times CLtree(3,q) + CLtree(1,q) \times 0.61 \times (1-CLtree(3,q))} &&
\end{flalign}
\begin{flalign}
	d_3 = \log(\frac{d3_{num}}{d3_{den}}) - d_q &&
\end{flalign}
\begin{flalign}
	d1^* = \frac{(1-CLtree(1,q)) \times 0.61 \times (1-CLtree(3,q))}{CLtree(1,q) \times 0.39 \times (1-CLtree(3,q)) + (1-CLtree(1,q)) \times 0.61 \times (1-CLtree(3,q))} &&
\end{flalign}
\begin{flalign}
	d_1 = \log(d1^*) &&
\end{flalign}
\begin{flalign}
	d4^* = \frac{CLtree(1,q) \times 0.61}{1- 0.39 \times CLtree(1,q)} &&
\end{flalign}
\begin{flalign}
	\label{equ16}
	d4 = \log(d4^*) - d_q &&
\end{flalign}
\begin{comment}
	\begin{equation}
		\label{equation7}
		A = \sum_{q=0}^{|\mathcal{C}|}[1-CLtree(1,q) + (1-newP)  \times [\frac{\alpha}{\alpha+\beta}]] \times(1-QimgDesc(0,q)>0)] \\
	\end{equation}
	\begin{multline}
		\label{equation8}
		B= \sum_{q=0}^{|\mathcal{C}|} [newP \times CLtree(1,q) + (1-newP)  \times 
		[1- \frac{\alpha}{\alpha+\beta}] \times  \\ CLtree(1,q) ] \times [(QimgDesc(0,q)>0) \times 0.39 + (1-(QimgDesc(0,q)>0) \times 0.61)]
	\end{multline}
	
\end{comment}
For each image, $q$ is the index of the word in the vocabulary, and ranges from $0$ to the $|\mathcal{C}|$. Filling confusion matrix can easily be achieved using the filled matches vector.
$d_3$ and $d_4$ are used to calculate using log-likelihood based on the content current observation. $d_2 $ is used to calculate log-likelihoods for unobserved words that are children of observed words in Chow-Liu tree.
In [\ref{equ7} - \ref{equ16}], CLtree refers to Chow-Liu tree obtained from the BoW representaion of images and it is constructed using the deep CAE. For more information about Chow-Liu tree, see appendix $B$ in \citep{mark2009a}:
\begin{comment}
	For each image, $q$ is the index of the word in the vocabulary, and ranges from $0$ to the $|\mathcal{C}|$, where $|\mathcal{C}|$ is obtained from applying an autoencoder on the extracted SURF feature. The procedure for filling confusion matrix for every image is described in steps $a$ through $d$, where $A|_{newP=0} + B|_{newP=0}$ is used to compute new-place likelihood. In \ref{equation7} and \ref{equation8}, CLtree refers to Chow-Liu tree obtained from the training vocabulary. For more information, see appendix $B$ in \citep{mark2009a}:
	%[label=\alph*)] \large 
\end{comment}
\begin{comment}	
	\begin{enumerate}[label=\alph*)] \large 
		\item $\forall i \in DescImage(l), Qmatch(0)=(i,A|_{newP=0}+B|_{newP=0})$
		\item $\forall j \in(1 ,|BoW|), Qmatch(j)=(j,A+B)$
		\item $\forall k \in(1 ,|Qmatch|), Qmatch(k).index = i$
		\item $matches = [matches, Qmatch]$
	\end{enumerate}
	
	Where $\alpha$ and $\beta$ are defined as follows:
	\begin{itemize}
		\label{one}
		\item $\beta = CLtree<1,q>$\\ 
		\item $\alpha = BoW[i][q] \times  0.39 + (1-BoW[i][q]) \times (0.69) \times CLtree<1,q>$
	\end{itemize}

	$A+B$ is the likelihood of being in a new-place, provided the Boolean variable $newP$ is false. We need $A + B$ to calculate likelihood, and steps $a$ trough $d$ are repeated on every image's descriptors. The match probability (normalized over other comparison likelihoods) constitutes elements of the confusion matrix. Also BoW is obtained from clustering centroids, which is obtained by applying an Autoencoder on SURF feature space.\\
\end{comment}
\subsection{AE-ORB-SLAM}
\begin{figure*}[!htp]
	\centering
	\begin{subfigure}[b]{\textwidth}
		\includegraphics[width=\textwidth]{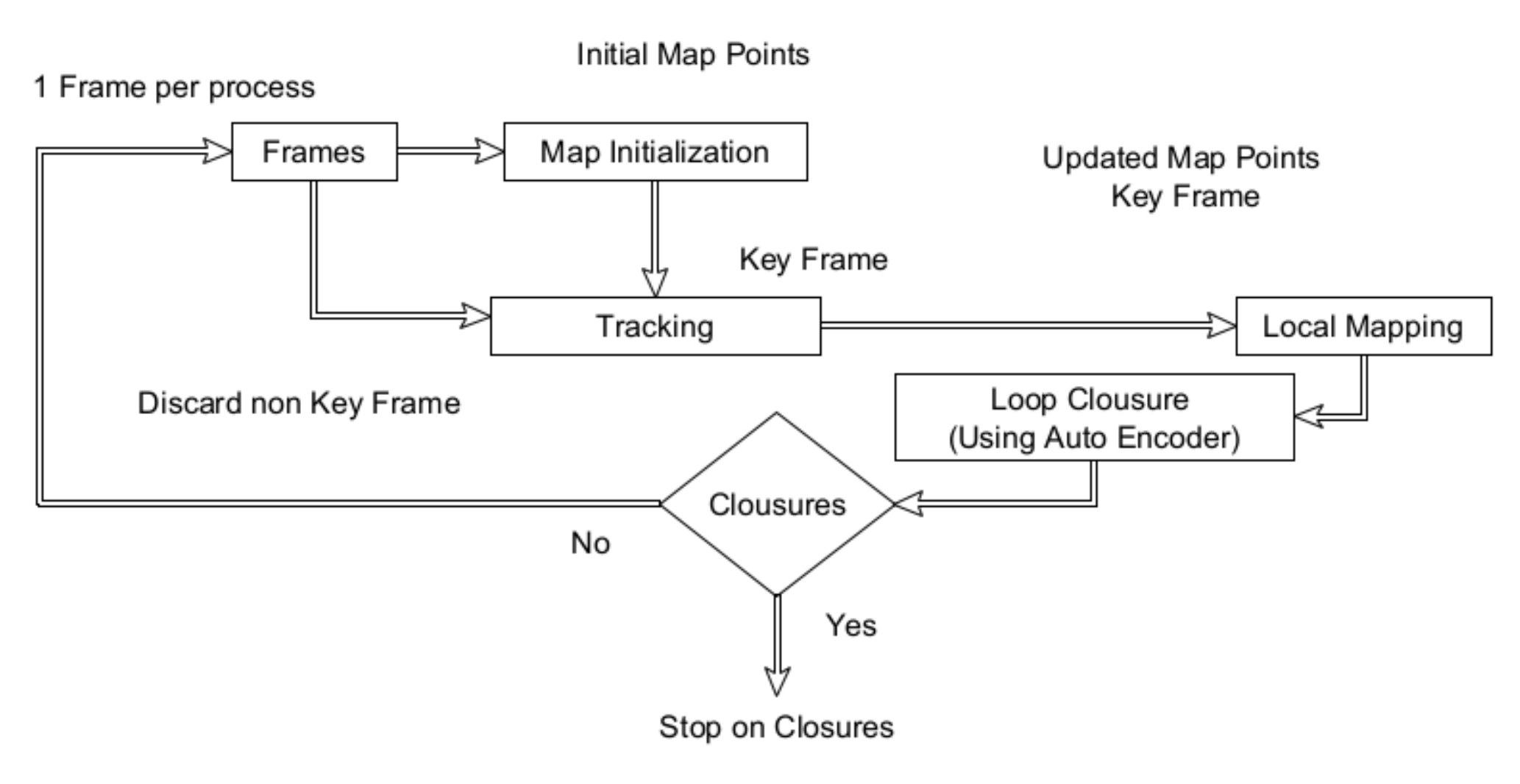}
		\label{fig:gull}
	\end{subfigure}
	\caption{Pipeline of Deep Convolutional AE-ORB SLAM}\label{fig 9}
\end{figure*}

As shown in pipeline of Figure \ref{fig 9}, the AE-ORB SLAM procedure is roughly the same as ORB-SLAM \cite{DBLP:journals/corr/Mur-ArtalMT15}, the only difference between these two SLAM algorithms is that in ORB-SLAM loop closure detection is computed by an unsupervised Kmeans clustering algorithm, while in AE-ORB-SLAM we replace that module with a self-supervised deep convolutional autoencoder. Furthermore, AE-ORB-SLAM  starts with an initialization of the points in the map, in the 3D space. Next, 3D map points and relative camera position  are computed  using the extracted ORB features trough triangulation.
In the tracking part, ORB-Features are matched with each other. from first frame to the last, and the refined estimated camera pose is returned. In the local mapping, if an image is detected to be a key-frame, a 3D map point corresponding to that frame is placed in the 3D map, this is achieved using bundle adjustment. In the loop closure detection part, as mentioned before the BoW representation of images is  constructed in the offline phase using a  deep convolutional autoencoder. \\ \\

\subsection{AE-BoW}
In regular BoW, the bag of words representation of images is obtained from cluster centroids. These centroids are  the result of applying Kmeans  to the standard image features. Furthermore, loop closure is detected by obtaining the cosine similarity of each image with all other images. In  AE-BoW, we repeat the same procedure, except that instead of Kmeans, we embed a deep convolutional  autoencoder to obtain the cluster centroids. This significantly increases the  performance LCD.
\section{Experimental Results}
\subsection{Metrics}

In this article we have used the metrics used in \citep{zhong}. It uses recall and accuracy of loop closure detection, They are defined as follows:\\

\begin{equation}
	\label{equation9}
	recall  =\mathlarger{\frac{\sum\limits_{i} \sum\limits_{j}{((Confusion \hspace{0.1cm} Matrix[i][j] \hspace{0.1cm} > threshold)\hspace{0.1cm} \wedge ground \hspace{0.1cm} truth[i][j]==1) }}{\sum\limits_{i} \sum\limits_{j}{ground \hspace{0.1cm} truth[i][j]==1}}}
\end{equation}

Each entry $(i^{th},j^{th})$ of the ground truth matrix  indicates whether images $i$ and $j$ were taken from the same location, if so, the entry is 1, otherwise it is zero.
True Positives, the numerator of formula ~\ref{equation9} and ~\ref{equation10}, is the number of elements in the confusion matrix greater than a given threshold, if their corresponding  elements in the ground truth matrix are also ones.
\\

Accuracy is defined as below:\\

\begin{equation}
	\label{equation10}
	accuracy  =\mathlarger{\frac{\sum\limits_{i} \sum\limits_{j}{((Confusion \hspace{0.1cm} Matrix[i][j] \hspace{0.1cm} > threshold)\hspace{0.1cm} \wedge ground \hspace{0.1cm} truth[i][j]==1) }}{\sum\limits_{i} \sum\limits_{j}{(Confusion \hspace{0.1cm} Matrix[i][j]==1  \hspace{0.1cm} > threshold)}}}
\end{equation}

\begin{comment}
	\begin{equation}
		\label{equation11}
		\mathcal{C}_i = \left\{
		\begin{array}{@{}l@{\thinspace}l}
			\text{if i=1  true  positive $\geq$ threshold}  & \text{$\wedge$ ground  thruth = 1  } \\
			\text{if i=2  true  positive $\leq$ threshold}  & \text{$\wedge$ ground  thruth = 0 }\\      
			\text{if i=3   true  positive $\geq$ threshold}  & \text{$\wedge$ ground  thruth = 0  }\\
			\text{if i=4   true  positive $\leq$ threshold}  & \text{$\wedge$ ground  thruth = 1  }\\
		\end{array}
		\right.
	\end{equation}
\end{comment}

\subsection{Datasets}
The methods presented in this paper are tested on several indoor and outdoor datasets. The four  methods BoW-SLAM, AE-Bow, FABMAP2 and AE-FABMAP are applied on Lip6 indoor dataset, TUM sequence 11, Stlucia (train/test), sequence 6 and 7 from Kitti dataset \citep{Geige}. For ORB-SLAM  the experiment is performed on TUM  Freiburg3  long office household. In the following section we briefly review the datasets used in this article. AE-FABMAP implementation is an extension to the  the C++ package provided by \citep{Glover}.\\

\subsubsection{Datasets used in AE-FABMAP and AE-BoW}
\textbf{Lip6indoor dataset\footnote[1]{https//animatlab.lip6.fr/AngeliVideosEn}}:  Contains 387, 240 $\times$ 192 images from a lab environment and contains several loops.

%is used as our testing dataset where it's  training data is monocular Technical university of  Munich sequence $11^{th}$

\textbf{TUM sequence 11\footnote[2]{https://vision.in.tum.de/data/datasets/mono-dataset}}: Contains 1500 images from a lab environment with different lighting condition. This dataset is collected from a narrow environment.\\

\textbf{New College [Cummins and Newman 2008]}: This dataset  has been taken from the Oxford university campus which includes complex repetitive structures. This is an stereo dataset with left and right sequence. We have used the left sequence containing 1073  with resolution of 640×480.\\

\textbf{Newer College}: Originally this dataset is a video (which we have converted into 200 image sequence). This datast can be considered as a portion  of new college dataset,  it also has been collected at night and it  contains 3 loops. In this dataset camera is experiencing cluttered movements.\\

\textbf{Stlucia suburbs}: There are two training and testing movies available for this dataset. Training data converted it to 540 images, and test dataset is converted to approximately 1000 images. This data set has been collected form saint Lucia suburbs. The environment in training and testing are roughly the same. \\

\subsubsection{Dataset used  in AE-ORB-SLAM}
\textbf{Freiburg$3$ \footnote[3]{https://vision.in.tum.de/rgbd/dataset/freiburg3/} }: This indoor dataset contains 2500 images taken from  area around a table, 
and it contains one loop.\\

\begin{figure*}[!htp]
	\centering
	\begin{subfigure}[b]{\textwidth}
		\includegraphics[width=\textwidth]{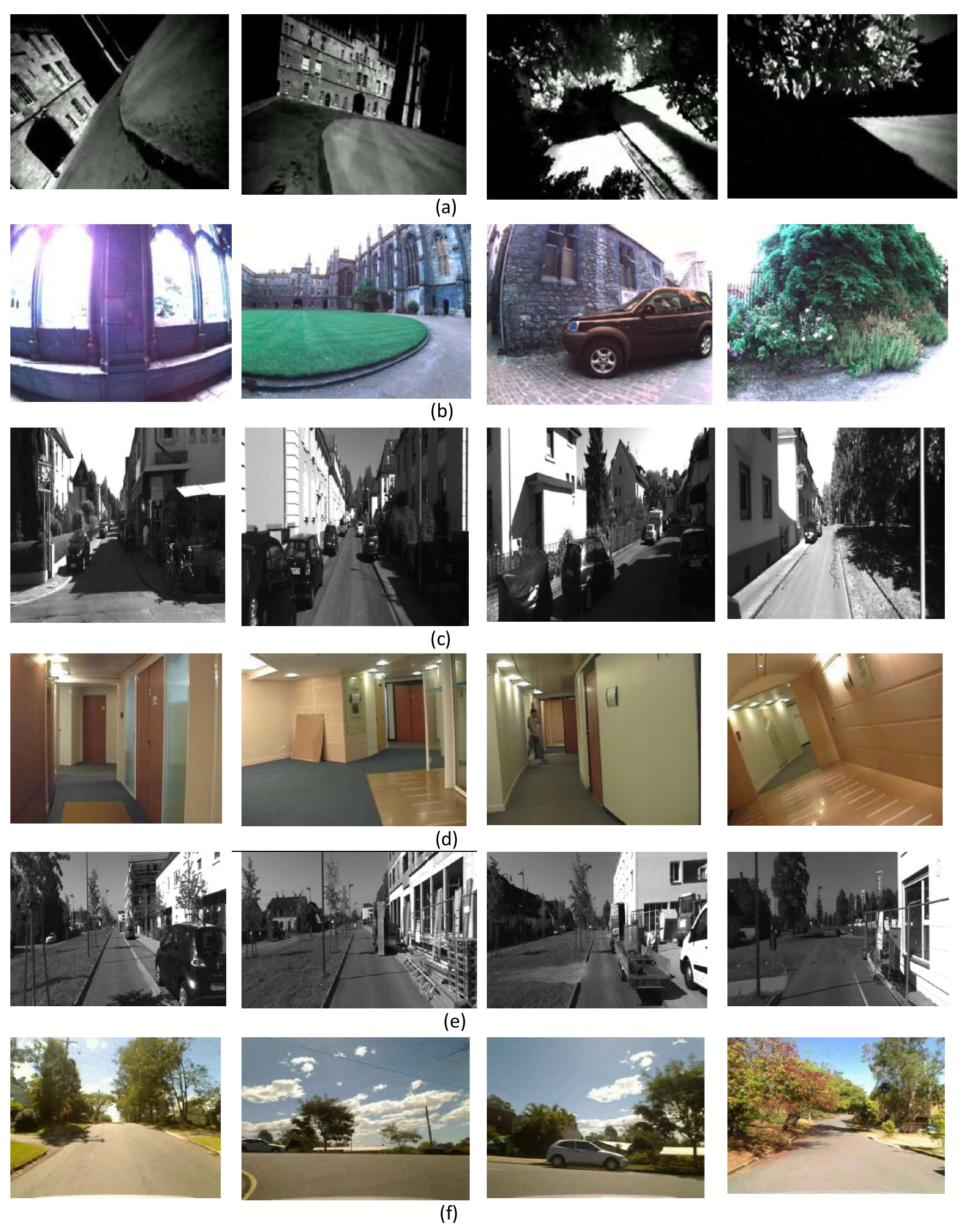}
	\end{subfigure}
	\caption{Datasets (a) Newer college, (b) New college, (c) Kitti sequence 7, (d) Lip6indoor, (e) Kitti sequence 6, (f) Stlucia (Tarin) }
	\label{fig 10}
\end{figure*}

\subsection{Results}

\begin{table}
	\centering
	\caption{Comparing  AE-FABMAP and FABMAP2  methods for LCD.}
	\label{ta}
	\begin{tabular}{|l|l|l|l|l|l|} 
		\hline
		Train Dataset                                                & Test Dataset                                                                              & \multicolumn{2}{l|}{AE-FABMAP}                                                                                                                                                               & \multicolumn{2}{l|}{FABMAP2}                                                                                   \\ 
		\hline
		&                                                                                           & \%Acc                                                                                       & \%Rec                                                                                          & \%Acc                                                  & \%Rec                                                 \\ 
		\hline
		\begin{tabular}[c]{@{}l@{}}Stlucia\\311417\end{tabular}      & \begin{tabular}[c]{@{}l@{}}Stlucia\textcolor[rgb]{0.502,0,0}{~ ~ ~}\\ 111417\end{tabular} & \begin{tabular}[c]{@{}l@{}}\%\textbf{74}\\ \% \textbf{83.5}\end{tabular}                    & \begin{tabular}[c]{@{}l@{}}\% \textbf{79.12}\\ \% \textbf{69.5}\end{tabular}                   & \%59.55                                                & \%81.5                                                \\ 
		\hline
		\begin{tabular}[c]{@{}l@{}}TUM seq11\\~ 3924499\end{tabular} & \begin{tabular}[c]{@{}l@{}}\\Lip6Indoor~ ~~\\~ 27189\end{tabular}                         & \begin{tabular}[c]{@{}l@{}}\%60.8\\\textbf{\%59.55}\end{tabular}                            & \begin{tabular}[c]{@{}l@{}}\%72.2\\\textbf{\%81.5}\end{tabular}                                & \%52.59                                                & \%81.2                                                \\ 
		\hline
		\begin{tabular}[c]{@{}l@{}}Kitti seq 6\\227904\end{tabular}  & \begin{tabular}[c]{@{}l@{}}Kitti seq 7\\235510\end{tabular}                               & \begin{tabular}[c]{@{}l@{}}\% \textbf{54.8}\\\% \textbf{60}\\ \% \textbf{63.9}\end{tabular} & \begin{tabular}[c]{@{}l@{}}\% \textbf{90.8}\\ \% \textbf{81.5}\\ \% \textbf{72.2}\end{tabular} & \%51.1                                                 & \%53                                                  \\ 
		\hline
		\begin{tabular}[c]{@{}l@{}}New College\\160500\end{tabular}  & \begin{tabular}[c]{@{}l@{}}Newer College\\194000\end{tabular}                             & \begin{tabular}[c]{@{}l@{}}\textbf{\%64.8}\\\textbf{\%80.2}\end{tabular}                    & \begin{tabular}[c]{@{}l@{}}\textbf{\%94.3}\\\%\textbf{87.4}\end{tabular}                       & \begin{tabular}[c]{@{}l@{}}\%60.1\\\%74.4\end{tabular} & \begin{tabular}[c]{@{}l@{}}\%89\\\%71.6\end{tabular}  \\
		\hline
	\end{tabular}
\end{table}

\begin{table}
	\centering
	\caption{Comparing AE-BoW and BoW  methods for LCD.}
	\label{ta1}
	\begin{tabular}{|l|l|l|l|l|l|} 
		\hline
		Train Dataset                                               & Test Dataset                                                                           & \multicolumn{2}{l|}{~ ~ ~ AE-BoW}                                                                                          & \multicolumn{2}{l|}{BoW}  \\ 
		\hline
		&                                                                                        & \%Acc                                                       & \%Rec                                                        & \%Acc  & \%ReC            \\ 
		\hline
		\begin{tabular}[c]{@{}l@{}}Stlucia~~\\~311417\end{tabular}  & \begin{tabular}[c]{@{}l@{}}Stlucia \textcolor[rgb]{0.502,0,0}{~}\\ 111417\end{tabular} & \begin{tabular}[c]{@{}l@{}}\\\%\textbf{83.8}\\\end{tabular} & \begin{tabular}[c]{@{}l@{}}\\\% \textbf{61.6}\\\end{tabular} & \%83.6 & \%60             \\ 
		\hline
		\begin{tabular}[c]{@{}l@{}}TUM seq11~\\3924499\end{tabular} & \begin{tabular}[c]{@{}l@{}}Lip6Indoor~~\\~ 27189\end{tabular}                          & \textbf{\%82.4}                                             & \textbf{\%82.1}                                              & \%46   & \%80             \\ 
		\hline
		\begin{tabular}[c]{@{}l@{}}Kitti seq 6\\227904\end{tabular} & \begin{tabular}[c]{@{}l@{}}Kitti seq 7\\235510\end{tabular}                            & \textbf{\%60}                                               & \textbf{\%75.6}                                              & \%45   & \%71             \\ 
		\hline
		\begin{tabular}[c]{@{}l@{}}New College\\160500\end{tabular} & \begin{tabular}[c]{@{}l@{}}Newer College~\\194000\end{tabular}                         & \textbf{\%71}                                               & \textbf{\%87}                                                & \%67   & \%87             \\
		\hline
	\end{tabular}
\end{table}

Table ~\ref{ta} compares  AE-FABMAP and FABMAP2 methods for the training datasets Saint Lucia(training), TUM sequence 11, Kitti sequence 6, New College  and testing datasets Saint Lucia(testing), lip6indoor, Kitti sequence 7 and Newer College, respectively. The results are only reported for testing datasets in term of accuracy and recall. With the same datasets the experiment is repeated for algorithms AE-BoW and BoW, and the result of comparison is presented in Table ~\ref{ta1}. In these experiments, training and testing datasets are chosen in a way that they share a similar content but without any repeated path. \\

Next, we compare ORB-SLAM method with the newly introduced method AE-ORB-SLAM. Here we are outperforming ORB-SLAM algorithm, still the results are competitive. Absolute RMSE for key frame trajectory (m) for  ORB-SLAM is  0.054353 and for our method it is 0.0473.
To implement deep AE-ORB-SLAM  we have modified the MATLAB ORB-SLAM package. \footnote[4]{https://www.mathworks.com/help/vision/ug/monocular-visual-simultaneous-localization-and-mapping.html}\\

\begin{figure*}[!htp]
	\centering
	\begin{subfigure}[b]{\textwidth}
		\includegraphics[width=\textwidth]{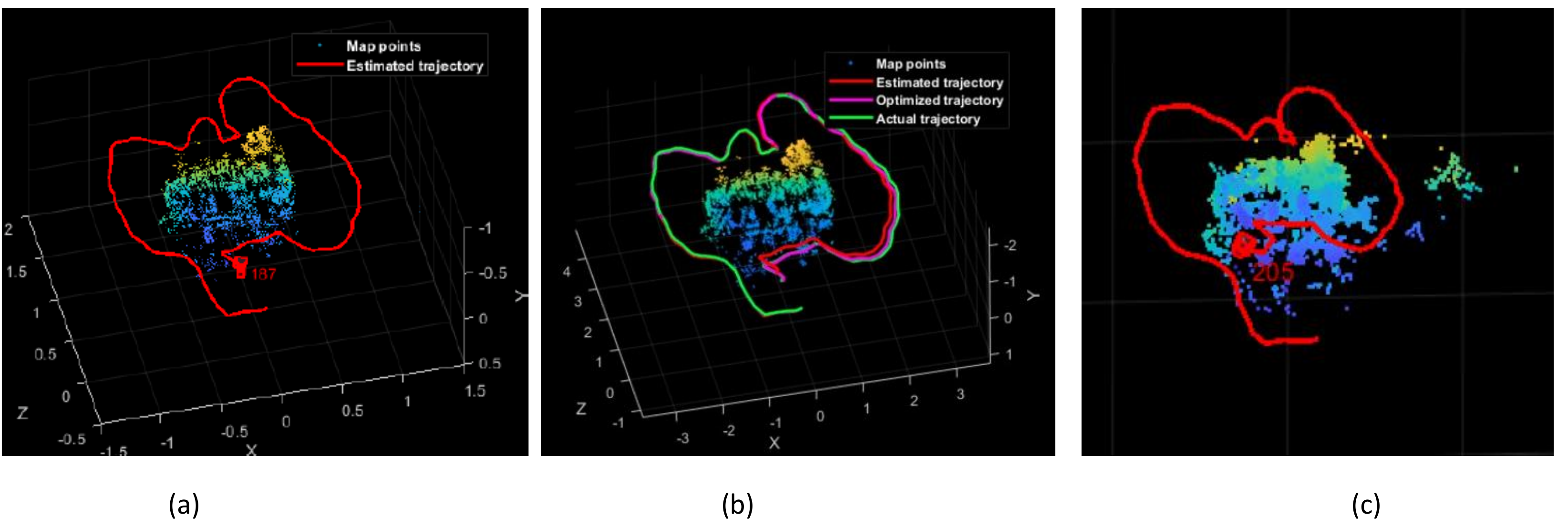}
		\label{fig:gull}
	\end{subfigure}
	\caption{Trajectory of Frieburg3 dataset using (a) ORB-SLAM, (b) Ground truth and (c)  AE-ORB-SLAM}\label{fig 11}
\end{figure*}

The trajectory obtained obtained from applying algorithms ORB-SLAM and AE-ORB-SLAM are shown in  parts (a) and (c) of figure \ref{fig 11} respectively. In this figure ground truth is plotted in section (b).    

\newpage
\begin{figure*}[!htp]
\begin{subfigure}[b]{\textwidth}
	\includegraphics[width=1\textwidth]{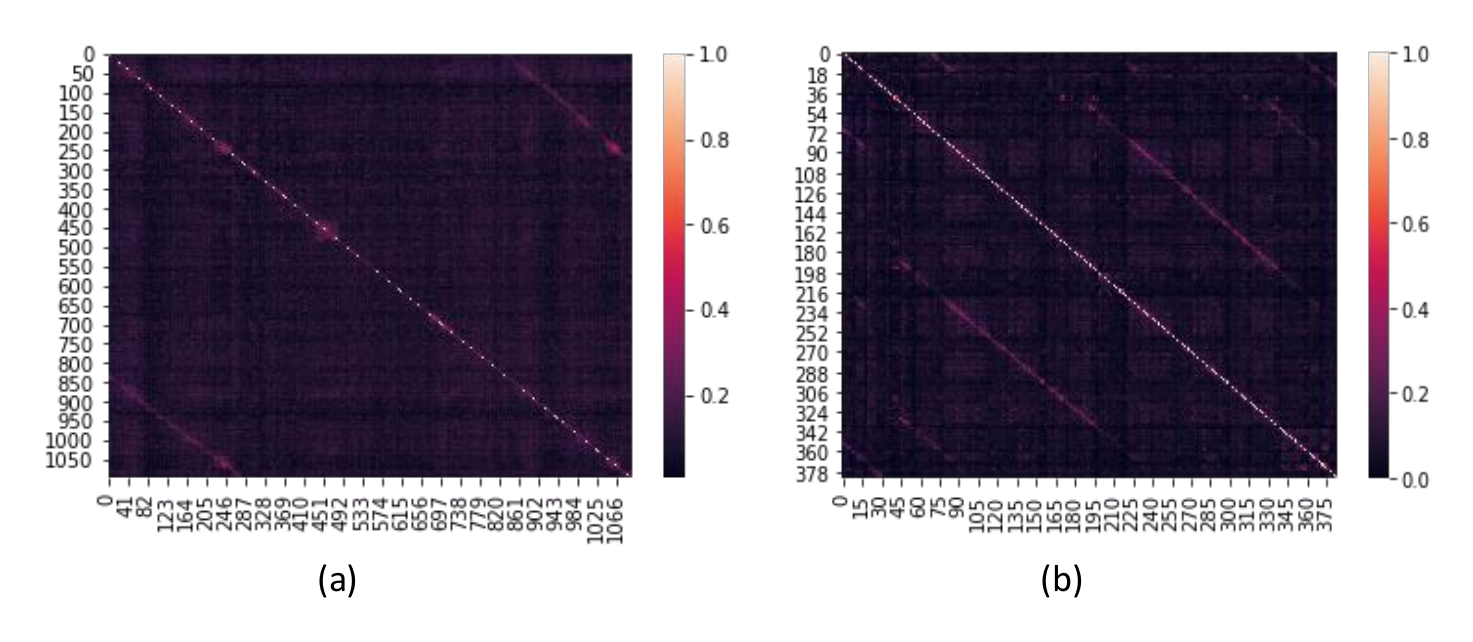}\\
	\includegraphics[width=1\textwidth]{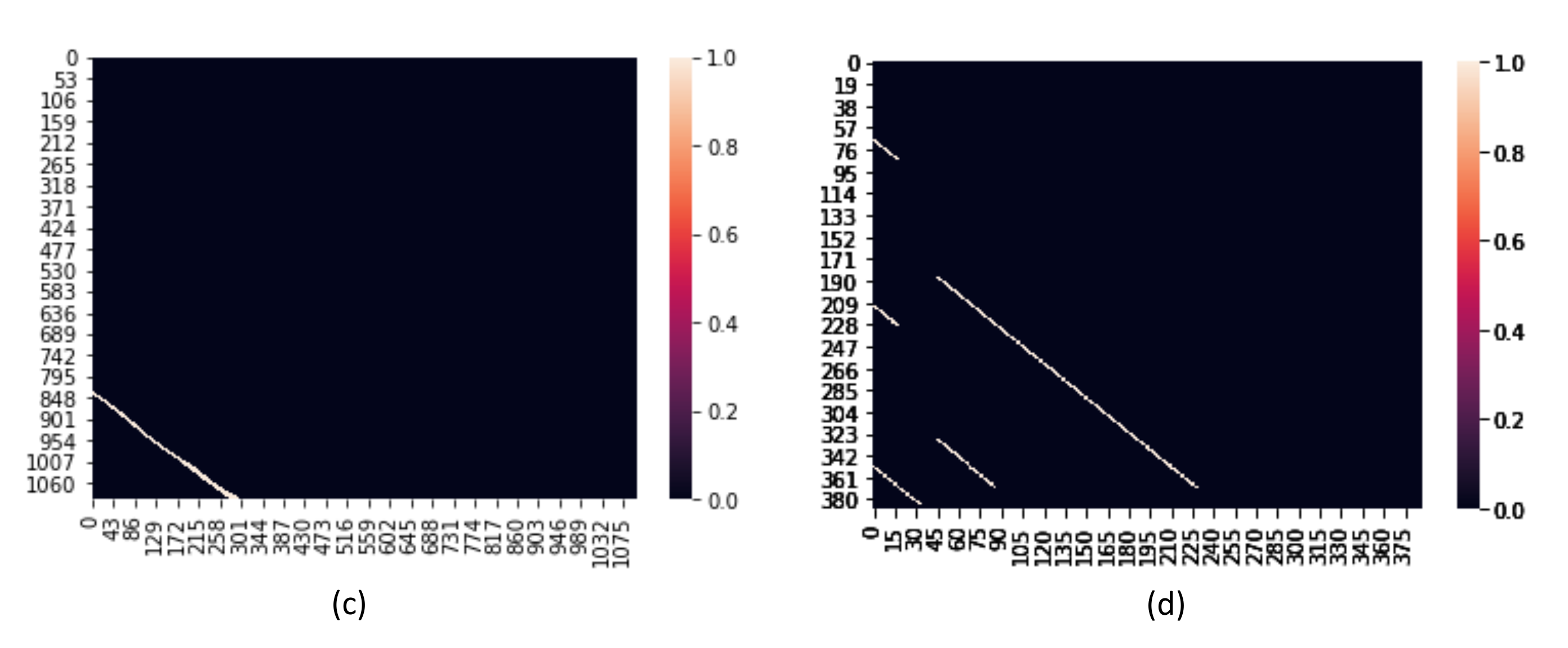}
	\label{fig:gull}
\end{subfigure}
\caption{(a) Confusion matrix for Kitti dataset (sequence 6), (b) Confusion matrix for ip6indoor dataset, (c) Ground truth
	for Kitti dataset, and (d) Ground truth for ip6indoor dataset.}\label{fig 12}
\end{figure*}

At the end of this section, for the sake of visual inspection of loop closures,  we have plotted the result of  AE-BoW (Due to it's fuzzy nature) applied on Lip6indoor dataset and Kitti sequence 6 dataset. These results are shown in figure~\ref{fig 12} in terms of confusion matrices. The resulting confusion matrices are
consistent  with the datasets ground truths.

\section{Conclusion}

In this paper we introduced extensions to the current state of the art SLAM methods,
and compared them with FABMAP2, ORB-SLAM and BoW. We concluded that the result of vector quantization module of the SLAM, may significantly enhance
loop closure detection  accuracy and recall. Our next task is to extend AE-
FABMAP to operate on long range. This is  achievable using deep autoencoders
because as  number of images increases the number of features also grows, but the problem   remains tractable using autocoders due to their efficiency in memory consumption. However we can apply autoencoders locally and join the clustering result using semi-supervised Latent Dirichlet Allocation,
(LDA), to form the global clusters.

\nocite{*}
\bibliographystyle{ieeetr}
\bibliography{ms}

\end{document}

%% file: elksty.tex
\def\E{\ifmmode{\mathbb E}\else{$\mathbb E$}\fi} %natural numbers
\def\N{\ifmmode{\mathbb N}\else{$\mathbb N$}\fi} %natural numbers
\def\R{\ifmmode{\mathbb R}\else{$\mathbb R$}\fi} %real numbers
\def\Q{\ifmmode{\mathbb Q}\else{$\mathbb Q$}\fi} %rational numbers
\def\C{\ifmmode{\mathbb C}\else{$\mathbb C$}\fi} %complex numbers
\def\H{\ifmmode{\mathbb H}\else{$\mathbb H$}\fi} %complex numbers
\def\Z{\ifmmode{\mathbb Z}\else{$\mathbb Z$}\fi} %integers
\def\P{\ifmmode{\mathbb P}\else{$\mathbb P$}\fi} %real numbers
\def\T{\ifmmode{\mathbb T}\else{$\mathbb T$}\fi} %real numbers
\def\SS{\ifmmode{\mathbb S}\else{$\mathbb S$}\fi} %real numbers
\def\DD{\ifmmode{\mathbb D}\else{$\mathbb D$}\fi} %real numbers

\newcommand{\bse}{\begin{subequations}}
\newcommand{\ese}{\end{subequations}}
\newcommand{\ben}{\begin{enumerate}}
\newcommand{\een}{\end{enumerate}}
\newcommand{\bens}{\begin{enumerate*}}
\newcommand{\eens}{\end{enumerate*}}
\newcommand{\be}{\begin{equation}}
\newcommand{\ee}{\end{equation}}
\newcommand{\bea}{\begin{eqnarray}}
\newcommand{\eea}{\end{eqnarray}}
\newcommand{\baa}{\begin{eqnarray*}}
\newcommand{\eaa}{\end{eqnarray*}}
\newcommand{\bc}{\begin{center}}
\newcommand{\ec}{\end{center}}

\theoremstyle{corollary}

\theoremstyle{lemma}

\theoremstyle{proposition}

\theoremstyle{axiom}

\theoremstyle{conjecture}

\theoremstyle{example}

\theoremstyle{definition}
\theoremstyle{remark}